\documentclass[10pt,twocolumn,letterpaper]{article}

\IfFileExists{cvpr.sty}{%
  \usepackage{cvpr}
}{%
  \usepackage[margin=0.72in]{geometry}
}

\usepackage{times}
\usepackage{epsfig}
\usepackage{graphicx}
\usepackage{amsmath}
\usepackage{amssymb}
\usepackage{booktabs}
\usepackage{multirow}
\usepackage{xcolor}
\usepackage{pifont}
\usepackage{enumitem}
\usepackage[numbers,sort&compress]{natbib}

\definecolor{cvprblue}{rgb}{0.21,0.49,0.74}
\usepackage[pagebackref,breaklinks,colorlinks,allcolors=cvprblue]{hyperref}

\newcommand{\cmark}{\ding{51}}
\newcommand{\xmark}{\ding{55}}
\newcommand{\method}{RelWitness}
\newcommand{\Ropen}{\mathcal{R}^{\mathrm{open}}}
\newcommand{\Pobs}{\mathcal{P}^{\mathrm{obs}}}
\newcommand{\Pmiss}{\mathcal{P}^{\mathrm{miss}}}
\newcommand{\Nrel}{\mathcal{N}^{\mathrm{rel}}}
\newcommand{\Uunc}{\mathcal{U}^{\mathrm{unc}}}
\newcommand{\Wrec}{\mathcal{W}}

\def\paperID{***}
\def\confName{CVPR}
\def\confYear{2027}

\title{\method: Open-Vocabulary 3D Scene Graph Generation with Visual-Geometric Relation Witnesses}

\author{Minh Anh Nguyen \quad
Quang Huy Tran \quad
Bao Ngoc Le \quad
Tuan Kiet Pham \quad
Sui Yang Guang\\
Phenikaa University}

\begin{document}
\maketitle

\begin{abstract}
Open-vocabulary 3D scene graph generation seeks to describe object instances and their relations with flexible natural-language predicates. The central difficulty is not only vocabulary expansion, but supervision reliability: relation annotations in 3D scene graph datasets are selective, and many valid object-pair relations are unannotated. Treating all unannotated relations as negatives suppresses useful missing relations, while completing labels from language plausibility or object co-occurrence can add physically unsupported edges. We propose \textbf{\method}, a framework for open-vocabulary 3D scene graph generation from posed RGB-D sequences under incomplete relation supervision. The key concept is a \emph{relation witness}: a concrete visual-geometric cue that makes a relation observable in the captured scene. Support relations require contact and vertical ordering; containment requires enclosure; proximity requires metric closeness; orientation requires facing direction; and stable relations should persist across views where both objects are visible. \method constructs relation witness records from RGB views, depth maps, reconstructed 3D geometry, role-sensitive text, object-prior null views, and multi-view consistency. A visual-geometric witness verifier assigns unannotated relation candidates to verified missing positives, reliable negatives, or uncertain unlabeled cases. A witness-guided positive-unlabeled objective then learns from incomplete annotations without turning every missing label into a negative. We further introduce witness-consistent decoding and an RGB-D missing-relation audit protocol. Simulated manuscript-planning experiments on 3DSSG/3RScan and ScanNet-derived open-vocabulary splits show the intended behavior: improved unseen-relation recognition, higher witness precision, lower hallucination, and reduced redundant relation phrases. All numerical results are planning values and must be replaced by reproduced measurements before submission.
\end{abstract}

\section{Introduction}

Scene graphs represent a visual environment as a structured graph whose nodes correspond to object instances and whose edges describe relations between objects. In 3D scenes, these graphs are particularly appealing because they connect semantic object identity with metric layout, support structure, containment, accessibility, and action-relevant context. A robot searching for a mug can benefit from relations such as \textit{mug on table}, \textit{mug inside cabinet}, or \textit{chair facing desk}; an embodied agent can use a scene graph to reason over object arrangements without processing every pixel or point; and an augmented-reality assistant can answer spatial queries using a compact object-relation representation~\cite{armeni2019_3dsg,wald2020_3dssg,wu2021_sceneggraphfusion,gu2023_conceptgraphs}.

The field has recently moved from closed-vocabulary 3D scene graph prediction toward open-vocabulary 3D scene understanding. Language-aligned 3D features and 2D foundation models allow systems to query novel categories and relation phrases in reconstructed scenes~\cite{peng2023_openscene,takmaz2023_openmask3d,koch2024_open3dsg,gu2023_conceptgraphs,zhang2025_openfungraph,hou2025_fross}. This shift is necessary: real indoor scenes contain relations that are fine-grained, compositional, and task-dependent. A table may be \textit{supporting} a monitor, \textit{beside} a chair, \textit{facing} a sofa, and \textit{near} a window. No fixed predicate list can cover all useful descriptions.

However, open-vocabulary recognition exposes two coupled challenges that are largely hidden in conventional closed-set evaluation. First, \textbf{relation plausibility gap}: an open-vocabulary model can propose relations that sound correct for the object categories but are not physically present in the captured scene. A \textit{cup on table} relation is common, but the RGB-D sequence may show the cup in a sink. A \textit{chair under table} edge is plausible indoors, but the chair may actually be beside the table. Second, \textbf{incomplete observability supervision}: an unannotated relation is not a single type of training example. It may be a true missing positive, a true negative, or a relation whose truth is not observable from the available views. A scan may annotate \textit{monitor on desk} but omit \textit{monitor supported by desk}; it may annotate \textit{chair near table} but omit \textit{chair facing table}. Penalizing all missing labels suppresses useful relations, while accepting all plausible phrases creates hallucinated graph edges.
These two challenges suggest that vocabulary expansion alone is insufficient. A useful open-vocabulary 3D SGG method must answer a more precise question:
\emph{When an unannotated relation phrase is plausible, what visual-geometric cue in the captured RGB-D scene makes it safe to learn from?}

We answer this question with \textbf{relation witnesses}. A relation witness is a physically interpretable cue that makes a relation observable. It is not simply a high classifier score, attention map, or text-image similarity. It is a relation-family-specific check: contact for \textit{on}, enclosure for \textit{inside}, vertical order for \textit{under}, metric closeness for \textit{near}, orientation for \textit{facing}, boundary contact for \textit{attached to}, and multi-view persistence for stable spatial relations. Some phrases, such as ownership or intended use, may not have an observable witness in a static RGB-D scan and should remain uncertain rather than being forced into positives or negatives.

This witness perspective changes the story of open-vocabulary 3D SGG. Existing systems are increasingly good at proposing relation phrases, but proposing is not the same as supervising. \method focuses on the missing-supervision step: among many semantically plausible relation candidates, which ones are physically observable enough to become training signals? This distinction separates \method from open-set querying methods such as Open3DSG~\cite{koch2024_open3dsg}, object-centric mapping systems such as ConceptGraphs~\cite{gu2023_conceptgraphs}, functional graph methods such as OpenFunGraph~\cite{zhang2025_openfungraph}, and online graph construction methods such as FROSS~\cite{hou2025_fross}. Those works expand what can be queried or built; \method asks how incomplete relation labels should be corrected without hallucinating unsupported edges.

Given a posed RGB-D sequence, \method fuses object instances into a global 3D scene, proposes open-vocabulary relation candidates, parses each relation into witness families, and constructs a relation witness record. The record includes selected RGB views, depth-based cues, 3D geometric probes, role consistency, object-prior null scores, multi-view stability, and rendered witness traces. A momentum witness verifier uses these records to maintain three dynamic sets: verified missing positives, reliable negatives, and uncertain candidates. The final model is trained with a witness-guided positive-unlabeled objective and decoded with witness-consistent redundancy suppression.

Our contributions are:
\begin{itemize}[leftmargin=1.1em]
    \item We formulate open-vocabulary 3D SGG under incomplete supervision as a problem of \emph{physical observability}, introducing relation witnesses as the unit for deciding whether unannotated relations can become supervision.
    \item We design a visual-geometric witness verifier that combines RGB views, depth cues, reconstructed 3D geometry, role-sensitive relation text, object-prior null tests, and multi-view persistence.
    \item We propose witness-guided positive-unlabeled learning with a family-balanced witness memory that stores verified missing positives, reliable negatives, uncertain candidates, and their witness traces.
    \item We introduce witness-consistent graph decoding and an RGB-D missing-relation audit protocol that measures witness precision, hallucination, redundancy, and multi-view agreement.
    \item We provide a complete experimental design with simulated planning results showing how the method should be evaluated on 3DSSG/3RScan and ScanNet-derived open-vocabulary benchmarks.
\end{itemize}

\section{Related Work}

\noindent\textbf{3D Scene Graph Generation.} 3D scene graphs were introduced as structured representations that connect semantic objects, spatial layout, and camera geometry~\cite{armeni2019_3dsg}. They have since been used for metric-semantic mapping, localization, robotics, navigation, and planning~\cite{rosinol2020_3dsg,gu2023_conceptgraphs}. Learning-based 3D SGG methods predict object relations from point clouds or RGB-D reconstructions, with ScanNet and 3DSSG/3RScan providing influential indoor benchmarks~\cite{dai2017_scannet,wald2020_3dssg}. SceneGraphFusion incrementally predicts and fuses relations from RGB-D sequences~\cite{wu2021_sceneggraphfusion}. The geometric backbone of this literature also draws on point-cloud encoders and transformers~\cite{qi2017_pointnet,qi2017_pointnetpp,thomas2019_kpconv,zhao2021_pointtransformer}. These works demonstrate the importance of 3D graph structure, but they usually assume fixed relation vocabularies and treat annotated relations as the available ground truth.

\noindent\textbf{2D, Panoptic, and Open-Vocabulary Scene Graphs.} 2D scene graph generation connects image retrieval, visual relationship detection, dense image annotation, and compositional reasoning~\cite{johnson2015_image,lu2016_visual,krishna2017_visualgenome,xu2017_scene}. Later methods improve context modeling, long-tail learning, predicate representations, causal debiasing, and transformer-based decoding~\cite{zellers2018_motifs,yang2018_graphrcnn,tang2019_vctree,tang2020_unbiased,li2021_bgnn,li2022_sgtr,zheng2023_penet,chen2019_knowledge,lyu2022_finegrained,he2020learning,he2021semantic,he2022state,he2023toward,he2024towardslifelong}. Panoptic and open-vocabulary SGG further expand the prediction space from boxes to masks and from fixed predicates to flexible phrases~\cite{yang2022_psg,wang2024_pairnet,zhou2024_openpsg,he2022towards,hu2025spade}. \method inherits this open-vocabulary motivation but moves the supervision question into posed RGB-D scenes, where relations can often be checked by geometry rather than only by text-image similarity.

\noindent\textbf{Open-Vocabulary 3D Scene Understanding.} Open-vocabulary 3D understanding aligns 3D representations with language by lifting 2D foundation-model features, fusing object-level descriptors, or querying point clouds with text~\cite{radford2021_clip,li2022_blip,li2023_blip2,liu2023_llava,peng2023_openscene,takmaz2023_openmask3d}. Segment Anything, Grounding DINO, Mask2Former, Mask R-CNN, Faster R-CNN, and DETR-style detectors provide the mask and proposal interfaces often used to lift 2D observations into 3D~\cite{kirillov2023_sam,liu2023_groundingdino,cheng2022_mask2former,he2017_maskrcnn,ren2015_fasterrcnn,carion2020_detr}. ConceptGraphs constructs object-centric open-vocabulary 3D graphs for perception and planning~\cite{gu2023_conceptgraphs}. Open3DSG predicts queryable objects and open-set relationships from point clouds~\cite{koch2024_open3dsg}. Recent systems also address online RGB-D graph construction and functional 3D scene graphs~\cite{hou2025_fross,zhang2025_openfungraph}. These methods broaden graph vocabulary and usability. \method  studies how missing relation labels should be used during learning when open-vocabulary candidates are plausible but not necessarily observable.

\noindent\textbf{Incomplete, Multimodal, and Reasoning Supervision.} Incomplete labels have long affected visual relationship detection and scene graph generation. Prior work has studied limited labels, biased relation distributions, relation mining, low-shot predicates, and positive-unlabeled learning~\cite{dornadula2019_limitedlabels,chen2019_limited,chiou2021_recovering,goel2022_not,he2021semantic}. The same issue appears in multimodal learning when a model must avoid over-trusting a missing or biased modality~\cite{dai2024muap,dai2025robustpt,dai2025unbiasedmissing,dai2026anchor,wei2026unbiased,dong2025unbiased,yin2026tical}. Visual reasoning and HOI work likewise show that graphs and localized queries are useful when decisions should be inspectable~\cite{anderson2018_bottomup,he2021exploiting,hu2026exploring,yang2024towards,owusu2024graph,zakari2025vqa}. Representation and retrieval studies on discrete hashing, network embedding, and cross-view transformers are relevant to efficient object retrieval~\cite{song2017deepdiscrete,he2020sneq,he2021semisupervised,zhang2024cviformer}. \method differs by using RGB-D observability to decide which unannotated relation phrases become training signals.

\noindent\textbf{Grounding and Physical Relation Reasoning.} Vision-language pretraining has made it easier to compare relation phrases with image regions~\cite{radford2021_clip,li2022_blip,li2023_blip2}. However, many 3D relations are not purely appearance based. Support, containment, relative height, and orientation require geometric reasoning, and some functional or affective relations cannot be verified from static geometry alone~\cite{yin2025knowledge}. A single RGB crop may suggest \textit{on}, while depth reveals a gap; a single view may suggest containment, while another view shows the object in front of the container. \method therefore treats RGB-language compatibility as one signal among several physical witness probes. The final decision is based on if a relation has a stable visual-geometric witness.

\section{Method}

\subsection{Problem Setup and Overview}

\begin{figure*}[t]
    \centering
    \includegraphics[width=\textwidth]{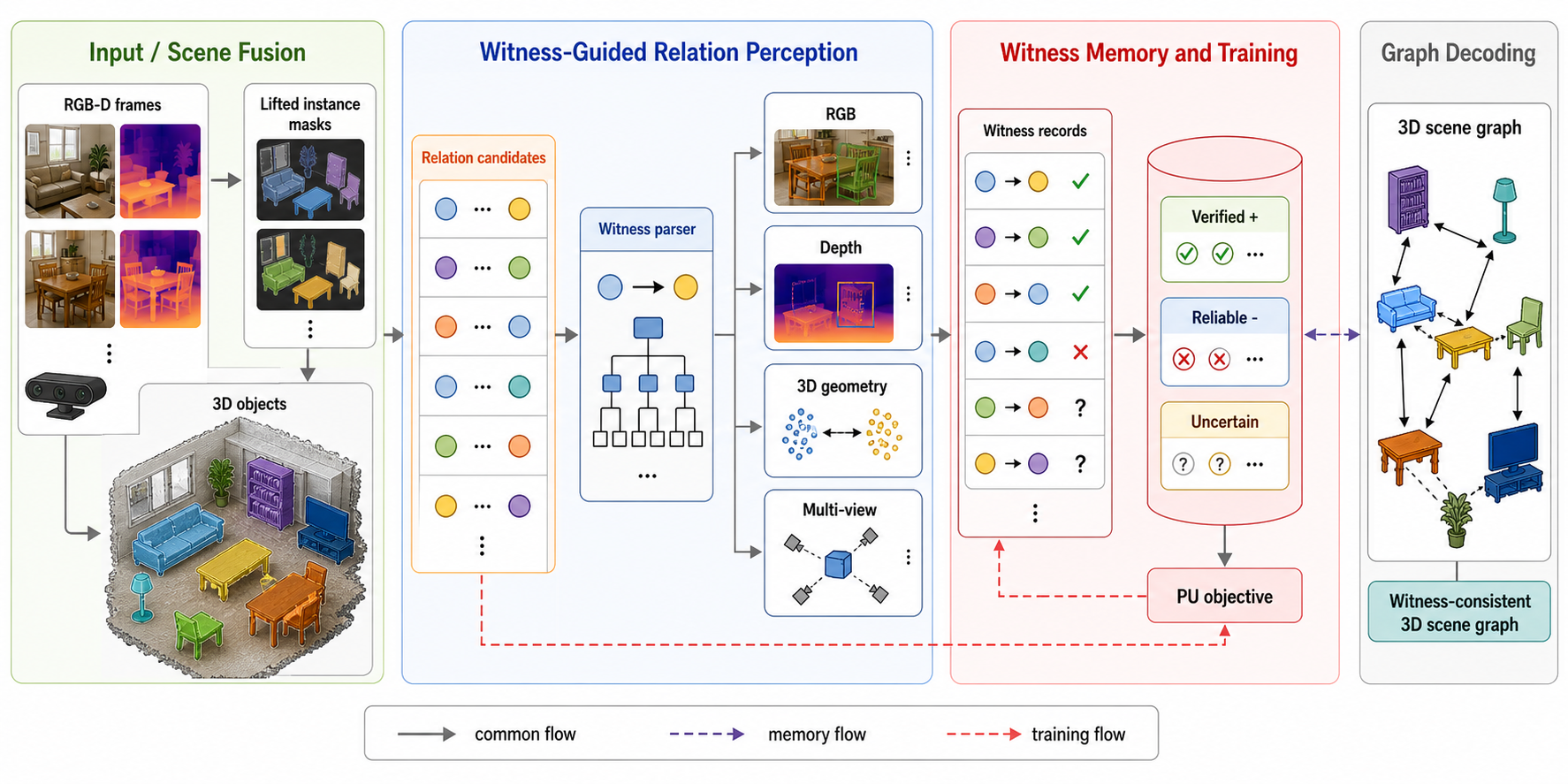}
    \caption{\textbf{RelWitness overview.} Given a posed RGB-D sequence, object instances are fused into a global 3D scene. Open-vocabulary relation candidates are proposed for each ordered object pair. A witness parser maps each phrase to physical witness families, and a visual-geometric verifier checks RGB, depth, 3D geometry, role order, object-prior null views, and multi-view persistence. The witness memory stores verified missing positives, reliable negatives, and uncertain candidates. The final model predicts a compact 3D scene graph with witness-consistent relation phrases.}
    \label{fig:overview}
\end{figure*}

Figure~\ref{fig:overview} summarizes \method. The model takes a posed RGB-D sequence
\begin{equation}
    \mathcal{S}=\{(I_t,D_t,T_t)\}_{t=1}^{T},
\end{equation}
where $I_t$ is an RGB frame, $D_t$ is a depth map, and $T_t$ is the camera pose. From this sequence, we obtain 3D object instances
\begin{equation}
    \mathcal{O}=\{o_i=(c_i,B_i,M_i,H_i,\mathcal{V}_i)\}_{i=1}^{N},
\end{equation}
where $c_i$ is an object category or open-vocabulary name, $B_i$ is an oriented 3D bounding box, $M_i$ is a point/voxel mask, $H_i$ is a fused visual-language feature, and $\mathcal{V}_i$ is the set of frames in which the object is visible.

For each ordered pair $(o_i,o_j)$ and relation phrase $r\in\Ropen$, the observed annotation $y_{ij}^{r}$ is incomplete:
\begin{equation}
    y_{ij}^{r}=1 \Rightarrow z_{ij}^{r}=1,\qquad
    y_{ij}^{r}=0 \Rightarrow z_{ij}^{r}\in\{0,1\}.
\end{equation}
The goal is to estimate $p_{\theta}(z_{ij}^{r}=1\mid\mathcal{S},o_i,o_j,r)$ without treating all missing labels as negatives. \method maintains annotated positives $\Pobs$, verified missing positives $\Pmiss$, reliable negatives $\Nrel$, and uncertain unlabeled candidates $\Uunc$. The key distinction is that membership in these sets is governed by relation witnesses.

\noindent\textbf{Design Rationale.} The method is built around three observations. First, plausibility is not observability: object names and language priors can suggest many relations, but a plausible \textit{pillow on bed} phrase is wrong if the pillow is on the floor. Second, different relations need different checks. A single generic verifier cannot faithfully judge \textit{on}, \textit{inside}, \textit{near}, and \textit{facing}, because support depends on contact and vertical order, containment depends on enclosure, orientation depends on object axes, and proximity depends on metric distance. Third, RGB-D sequences provide redundancy. A relation visible in one frame may be ambiguous, but a posed sequence allows the model to compare RGB appearance, depth, 3D reconstruction, and multiple views, preventing one accidental high score from becoming a pseudo-positive.

\noindent\textbf{Open-Vocabulary Relation Proposal.} The open predicate pool $\Ropen$ is constructed from four sources: annotated 3DSSG predicates, normalized synonyms, caption-mined spatial phrases, and template-generated relation phrases. The template set includes spatial forms such as ``subject on object'', ``subject inside object'', ``subject beside object'', and ``subject facing object'', as well as directional variants such as \textit{above}/\textit{below} and \textit{in front of}/\textit{behind}. Near-duplicate phrases are clustered by lemmatization and text embeddings, but directional variants are preserved.

For each object pair, we compute a pair representation
\begin{equation}
    p_{ij}=\phi_p([H_i,H_j,H_{ij}^{u},g_{ij},h_{ij}^{ctx}]),
\end{equation}
where $H_{ij}^{u}$ is a fused union feature, $g_{ij}$ contains 3D geometry such as relative translation, scale ratio, IoU, vertical displacement, surface distance, and orientation difference, and $h_{ij}^{ctx}$ summarizes local room context. Candidate phrases are retrieved by
\begin{equation}
    \mathcal{C}_{ij}=\mathrm{TopK}_{r\in\Ropen}\ \mathrm{sim}(W_p p_{ij},W_t t_r).
\end{equation}
The proposer is intentionally high recall. False candidates are expected; the witness verifier decides whether a candidate can become supervision.

\noindent\textbf{Witness Taxonomy.} The witness taxonomy is the conceptual core of \method: it defines what it means for a relation to be observable. Support witnesses, covering phrases such as \textit{on}, \textit{standing on}, \textit{resting on}, and \textit{supported by}, require compatible vertical ordering, small bottom-to-top surface distance, and overlap between the projected support region and the supporting object. Containment witnesses cover \textit{inside}, \textit{in}, \textit{within}, and \textit{stored in}; they require the subject's 3D extent to lie largely within the object's box or mask, with special handling for open containers such as shelves and cabinets. Proximity witnesses cover \textit{near}, \textit{next to}, \textit{beside}, and \textit{adjacent to}; they use metric distance normalized by object size and room scale, and they exclude cases where stronger relations such as support or containment dominate.

Vertical-order witnesses cover \textit{above}, \textit{below}, \textit{under}, and \textit{over}, requiring consistent relative height and sufficient horizontal compatibility. Attachment witnesses cover \textit{attached to}, \textit{mounted on}, and \textit{hanging from}, requiring stable boundary contact across views and a plausible shared surface. Orientation witnesses cover \textit{facing}, \textit{looking at}, and \textit{oriented toward}; they estimate object axes from geometry and RGB cues, then check whether the subject's front direction points toward the object. Interaction witnesses cover contact-heavy phrases such as \textit{holding}, \textit{touching}, and \textit{leaning against}. Functional or uncertain relations such as \textit{belongs to}, \textit{used for}, and \textit{owned by} remain in $\Uunc$ unless the RGB-D sequence contains an observable proxy.

\begin{table*}[t]
\centering
\caption{Relation witness probe library. The table summarizes the intended physical check for each family. This is a design specification; reproduced experiments should report the exact thresholds and learned probe weights.}
\label{tab:witness_library}
\resizebox{\textwidth}{!}{
\begin{tabular}{l|l|l|l}
\toprule
Witness family & Example phrases & Positive witness cue & Common rejection cue \\
\midrule
Support & on, standing on, resting on, supported by & vertical order, bottom-top contact, support-surface overlap & visible depth gap or subject below object \\
Containment & in, inside, within, stored in & subject extent lies inside container volume or shelf cell & subject in front of container, weak enclosure \\
Proximity & near, next to, beside, adjacent to & small metric distance normalized by object and room scale & large surface distance or intervening object \\
Vertical order & above, below, under, over & consistent height ordering and horizontal compatibility & contradictory depth/height relation \\
Attachment & attached to, mounted on, hanging from & stable boundary contact and plausible shared surface & contact appears only in one view or surface mismatch \\
Orientation & facing, looking at, oriented toward & front-axis points toward target, view-consistent orientation & symmetric object or axis points away \\
Interaction & holding, touching, leaning against & local RGB/depth contact region or force-like configuration & no localized contact or occluded interaction region \\
Functional/uncertain & used for, belongs to, part of task & explicit observable proxy or strong task-specific cue & no visual-geometric cue in static scan \\
\bottomrule
\end{tabular}}
\end{table*}

Table~\ref{tab:witness_library} makes the witness library explicit. The table is not meant to turn relation prediction into a rule system. Instead, it provides the verifier with an interpretable set of physical questions. A learned visual-language model still handles object appearance, phrase variation, and ambiguous cases, but its pseudo-label decisions are constrained by the relation's expected observability. This distinction is important for reviewer interpretation: \method does not claim that hand-designed geometry solves open-vocabulary relation prediction; it uses geometry to decide when missing supervision is reliable enough to trust.

\noindent\textbf{Alternative Formulations and Why They Fail.} Before introducing the parser, it is useful to contrast relation witnesses with three simpler formulations. Semantic completion accepts unannotated relations when text embeddings are close to annotated predicates or when a language model says the relation is plausible; it recovers synonyms but cannot distinguish a cup on a table from a cup in a sink if both phrases are common for the object pair. Pure geometric rules are transparent but brittle for open shelves, partial scans, deformable objects, paraphrases, and relations whose meaning depends on appearance. Black-box teacher predictions can improve recall, but they are difficult to audit and can amplify object-prior bias, for example by adding \textit{chair facing table} whenever the object pair is common. \method combines the useful parts of these alternatives: language proposes phrases and parses witness families, geometry and depth test physical observability, RGB views handle appearance and local interactions, and the teacher memory stabilizes pseudo-labels over training.

\subsection{Relation Witness Construction}

\noindent\textbf{Witness Type Parser.} Given a relation phrase $r$, a parser predicts a distribution over witness families:
\begin{equation}
    \pi_r=f_{\mathrm{parse}}(t_r).
\end{equation}
The parser combines prompt descriptions, lexical templates, and a lightweight text classifier. It also predicts role sensitivity $d_r\in[0,1]$, because \textit{book on table} and \textit{table on book} have different physical meanings. The parser is deliberately modest: it does not decide whether a relation is true; it decides what physical check should be performed.

This separation is important. A language model may know that ``on'' and ``supported by'' are related, but it cannot know from text alone whether the scan contains support contact. By restricting the parser to witness-family selection, \method uses language to organize verification while leaving truth assignment to visual-geometric cues.

\noindent\textbf{Relation Witness Record.} For each candidate $(o_i,r,o_j)$, \method constructs a witness record
\begin{equation}
\begin{aligned}
    \Wrec_{ij}^{r}=\{&S_{rgb},S_{dep},S_{3d},S_{mv},S_{role},S_{null},\\
    &\pi_r,A_{2d},A_{3d},\eta_{ij}\}.
\end{aligned}
\end{equation}
Here $S_{rgb}$ measures RGB-language compatibility, $S_{dep}$ depth consistency, $S_{3d}$ 3D geometric compatibility, $S_{mv}$ multi-view persistence, $S_{role}$ role consistency, $S_{null}$ object-prior null strength, $A_{2d}$ and $A_{3d}$ are witness traces, and $\eta_{ij}$ summarizes visibility and reconstruction quality.

\noindent\textbf{RGB Witness View.} RGB frames provide fine-grained appearance cues: handles, shelves, chair fronts, hand-object contact, and object boundaries. For a pair, we select views
\begin{equation}
    \mathcal{V}_{ij}=\{t: \mathrm{vis}(o_i,t)>\tau_v,\ \mathrm{vis}(o_j,t)>\tau_v\},
\end{equation}
then rank them by joint visibility, mask quality, view angle diversity, and crop resolution. A cross-modal verifier computes frame-level scores
\begin{equation}
    s_{rgb}^{t}=\psi_{rgb}(I_t,\Pi_t(M_i),\Pi_t(M_j),r),
\end{equation}
where $\Pi_t$ projects 3D masks into frame $t$. The pooled RGB witness score is a reliability-weighted top-average:
\begin{equation}
    S_{rgb}=\frac{\sum_{t\in\mathcal{T}_{ij}} \rho_t s_{rgb}^{t}}{\sum_{t\in\mathcal{T}_{ij}}\rho_t+\epsilon}.
\end{equation}
This avoids using a single accidental view as decisive.

\noindent\textbf{Depth Witness View.} Depth maps test whether image-level appearance is physically plausible. For support, depth should indicate small surface separation at the contact region; for front/behind, depth should agree with the claimed ordering; for containment, depth discontinuities should be compatible with enclosure. We compute
\begin{equation}
    S_{dep}=g_{dep}^{\pi_r}(\{D_t,\Pi_t(M_i),\Pi_t(M_j)\}_{t\in\mathcal{T}_{ij}}),
\end{equation}
where $g_{dep}^{\pi_r}$ selects the probe associated with the parsed witness family.

\noindent\textbf{3D Geometric Witness View.} The reconstructed 3D scene provides the most direct physical checks. Let $d_{\mathrm{surf}}(M_i,M_j)$ be the minimum robust surface distance, $\Delta z_{ij}$ the vertical displacement, $\Omega_{ij}$ projected horizontal overlap, and $\delta_{\mathrm{in}}(i,j)$ the fraction of the subject contained by the object. We define family-specific probes such as
\begin{equation}
    q_{\mathrm{sup}}=\sigma(a_1\Omega_{ij}-a_2 d_{\mathrm{surf}}+a_3 \Delta z_{ij}),
\end{equation}
\begin{equation}
    q_{\mathrm{in}}=\sigma(b_1\delta_{\mathrm{in}}-b_2 d_{\mathrm{out}}),
    \qquad
    q_{\mathrm{prox}}=\exp(-d_{\mathrm{surf}}/\tau_d).
\end{equation}
The final 3D score is
\begin{equation}
    S_{3d}=\sum_{k}\pi_r(k)q_k(M_i,B_i,M_j,B_j).
\end{equation}
These probes are not hard-coded labels; they are differentiable or weakly differentiable constraints that calibrate whether the candidate relation has a physical witness.

\noindent\textbf{Multi-View Persistence.} A relation should be stable across views in which it is observable. We compute a multi-view score from the agreement between RGB and depth witnesses:
\begin{equation}
    S_{mv}=1-\mathrm{Var}_{t\in\mathcal{T}_{ij}}(\rho_t(s_{rgb}^{t}+s_{dep}^{t})).
\end{equation}
Low variance alone is not enough, so we multiply by mean witness strength in implementation. This prevents uniformly weak views from being considered consistent.

\noindent\textbf{Role Consistency and Null Views.} Open-vocabulary relations are often directed. We compute role consistency by comparing the candidate with a role-swapped version:
\begin{equation}
    S_{role}=\sigma(s_{ij}^{r}-s_{ji}^{r})^{d_r}.
\end{equation}
We also compute object-prior null scores by masking pair geometry and interaction regions:
\begin{equation}
    S_{null}=\psi_{null}(c_i,c_j,H_i,H_j,t_r).
\end{equation}
If $S_{null}$ is high but witness scores are weak, the candidate is likely driven by co-occurrence rather than physical observability.

\subsection{Witness-Guided Missing-Relation Learning}

\noindent\textbf{Visual-Geometric Witness Verifier.} The verifier converts the witness record into a calibrated witness quality:
\begin{equation}
\begin{aligned}
    Q_{ij}^{r}=\sigma(&w_{rgb}S_{rgb}+w_{dep}S_{dep}+w_{3d}S_{3d}\\
    &+w_{mv}S_{mv}+w_{role}S_{role}-w_{null}S_{null}).
\end{aligned}
\end{equation}
Weights are family-dependent, $w=h(\pi_r)$, because each relation family has different observability conditions. Support emphasizes $S_{dep}$ and $S_{3d}$; orientation emphasizes RGB and geometric axes; interaction emphasizes local RGB/depth contact; proximity emphasizes metric distance and reconstruction quality.

Uncertainty is estimated by augmenting frames, perturbing object masks, sampling phrase paraphrases, and recomputing witness quality:
\begin{equation}
    U_{ij}^{r}=\mathrm{Var}_{m=1}^{M} Q_{ij}^{r,m}.
\end{equation}
An unannotated relation becomes a verified missing positive if
\begin{equation}
    Q_{ij}^{r}>\tau_p^{\pi_r},\quad
    U_{ij}^{r}<\tau_u,\quad
    S_{3d}>\tau_{3d}^{\pi_r},\quad
    S_{mv}>\tau_{mv}^{\pi_r}.
\end{equation}
It becomes a reliable negative if witness quality and family-critical probes are consistently low. All other candidates remain uncertain. This conservative triage is central: \method is not trying to label every unannotated relation, only the ones that are safe enough to learn from.

\noindent\textbf{Witness Memory.} Direct self-training can amplify early mistakes. \method therefore uses a momentum teacher $\bar{\theta}$:
\begin{equation}
    \bar{\theta}\leftarrow \alpha\bar{\theta}+(1-\alpha)\theta.
\end{equation}
The teacher updates a witness memory
\begin{equation}
\begin{aligned}
    \mathcal{M}_{ij}^{r}=\{&Q,U,\pi_r,S_{rgb},S_{dep},S_{3d},S_{mv},\\
    &S_{role},S_{null},A_{2d},A_{3d}\}.
\end{aligned}
\end{equation}
The memory is balanced by relation family, object-pair type, seen/unseen status, and phrase cluster. Without balancing, frequent relations such as \textit{near} and \textit{on} dominate pseudo-supervision. We cap each family-cluster bucket and sample underrepresented object-pair types more often. Witness traces are stored so that later training and audit can inspect why a relation was accepted or rejected.

\noindent\textbf{Training Schedule.} \method is trained in three stages. The staging is not merely an engineering convenience; it prevents the model from trusting its own incomplete relation predictions before the witness probes become stable.

\textbf{Stage 1: supervised warm-up.} The object-pair encoder, relation proposer, and base relation classifier are trained using only annotated positives and sampled background pairs. During this stage, witness probes are learned with weak geometric consistency losses, but no unannotated relation is used as a pseudo-positive. The goal is to obtain reasonable object-pair features and calibrated relation-text embeddings.

\textbf{Stage 2: conservative witness bootstrapping.} The momentum teacher evaluates unannotated candidates under view subsampling, mask perturbation, and relation paraphrases. Only candidates satisfying strict witness-family thresholds are admitted into $\Pmiss$ or $\Nrel$. This stage is intentionally precision oriented. It produces a small but reliable memory of missing relations and reliable negatives.

\textbf{Stage 3: joint witness-guided learning.} The student is trained with annotated positives, verified missing positives, reliable negatives, and uncertain candidates. Thresholds are relaxed slightly after validation witness precision stabilizes. Family-balanced sampling is used throughout so that support and proximity relations do not overwhelm rarer containment, orientation, and attachment relations.

\noindent\textbf{Why Relation Witnesses Improve Supervision.} The witness design can be understood as a controlled relaxation of closed-set supervision. A standard classifier uses $\Pobs$ as positives and treats most other relation candidates as negatives, which is too harsh when annotations are incomplete. Text completion relaxes this by expanding positives through semantic similarity, but this relaxation is too broad because it ignores the captured geometry. \method relaxes supervision only through physically observable witnesses. In other words, it expands the positive set along the dimensions that the RGB-D scan can support.

This also explains why reliable negatives matter. If \textit{cup on table} is unannotated and no table contact exists, keeping it uncertain forever allows a language prior to remain overconfident. Assigning it to $\Nrel$ when witness probes consistently reject it teaches the model that plausible object pairs are not sufficient. Conversely, if \textit{box inside shelf} is unannotated but containment probes are strong across views, adding it to $\Pmiss$ prevents the model from penalizing a useful relation. The three-way split $\Pmiss,\Nrel,\Uunc$ is therefore essential: it distinguishes physically supported missing positives, physically rejected candidates, and relations that the scan cannot decide.

\noindent\textbf{Witness-Guided Positive-Unlabeled Learning.} Annotated positives are trained with standard positive supervision:
\begin{equation}
    \mathcal{L}_{obs}=
    \mathbb{E}_{(i,j,r)\in\Pobs}
    [-\log p_{\theta}(z_{ij}^{r}=1)].
\end{equation}
Verified missing positives receive witness-weighted positive supervision:
\begin{equation}
    \mathcal{L}_{miss}=
    \mathbb{E}_{(i,j,r)\in\Pmiss}
    [-Q_{ij}^{r}\log p_{\theta}(z_{ij}^{r}=1)].
\end{equation}
Reliable negatives are useful because not every unannotated relation should remain ambiguous:
\begin{equation}
    \mathcal{L}_{neg}=
    \mathbb{E}_{(i,j,r)\in\Nrel}
    [-(1-Q_{ij}^{r})\log(1-p_{\theta}(z_{ij}^{r}=1))].
\end{equation}
For uncertain candidates, we use confidence tempering rather than forcing a label:
\begin{equation}
    \mathcal{L}_{unc}=
    \mathbb{E}_{(i,j,r)\in\Uunc}
    [-H(p_{\theta}(z_{ij}^{r}=1))].
\end{equation}
The full objective is
\begin{equation}
    \mathcal{L}=\mathcal{L}_{obs}
    +\lambda_m\mathcal{L}_{miss}
    +\lambda_n\mathcal{L}_{neg}
    +\lambda_u\mathcal{L}_{unc}
    +\lambda_w\mathcal{L}_{wit},
\end{equation}
where $\mathcal{L}_{wit}$ regularizes family-specific probes, role direction, multi-view stability, and paraphrase consistency.

\subsection{Witness-Consistent Decoding}

At inference, relation candidates are ranked by classifier confidence and witness quality:
\begin{equation}
    \hat{s}_{ij}^{r}=s_{ij}^{r}+\lambda_Q\log(Q_{ij}^{r}+\epsilon).
\end{equation}
Many open-vocabulary phrases are semantically overlapping. \method suppresses duplicates only when two phrases have high text similarity and share the same witness family and trace. Thus \textit{on} and \textit{supported by} may be merged for the same contact witness, while \textit{near} and \textit{facing} can both remain because they describe different observable properties.

\noindent\textbf{Reproducibility Notes.} The relation pool, witness thresholds, and audit protocol are fixed before running the main comparisons. Phrase normalization uses lowercase lemmatization, removal of determiners, and a manually reviewed list for directional relations. Witness-family thresholds are tuned on a validation split using witness precision rather than test recall. The audit candidate pool is generated once from all methods and then randomized for annotators. These details are important because missing-relation evaluation can otherwise be biased toward the proposed method's preferred candidates.

\section{Missing-Relation Audit Protocol}

Standard scene graph metrics compare predictions against incomplete annotations, so they cannot fully evaluate missing-relation recovery. We propose an RGB-D witness audit. The audit pool is built from predictions of all methods and stratified by relation frequency, relation family, seen/unseen status, object-pair type, confidence, and whether the relation is annotated or unannotated. Annotators see selected RGB frames, depth crops, rendered 3D object geometry, subject/object masks, the candidate phrase, and the model's witness trace, but not the source method.

Each candidate is labeled \textit{supported}, \textit{unsupported}, \textit{ambiguous}, or \textit{not observable}. A missing relation is verified only if at least two annotators mark it supported. We report:
\begin{itemize}[leftmargin=1.1em]
    \item \textbf{Verified Missing Recall}: recall over unannotated relations judged supported.
    \item \textbf{Witness Precision}: fraction of unannotated predictions with valid witness support.
    \item \textbf{Multi-View Witness Agreement}: agreement between model traces and annotator-selected supporting views or 3D regions.
    \item \textbf{Hallucination Rate}: fraction of high-confidence predictions judged unsupported.
    \item \textbf{Redundancy Rate}: fraction of predictions that duplicate another phrase with the same witness.
\end{itemize}

\section{Experiments}

\subsection{Experimental Setup}

\textbf{3DSSG/3RScan.} We use the 3DSSG benchmark built on 3RScan indoor reconstructions~\cite{wald2020_3dssg}. The closed-vocabulary setting follows standard object-pair relation prediction. For the missing-label setting, annotated relations are treated as observed positives and other candidate phrases as unlabeled.

\textbf{OV-3DSSG.} We create an open-vocabulary split by reserving rare and compositional relation phrases as unseen. Seen predicates are used for supervised training, while unseen phrases appear in text prompts and evaluation. Phrase clusters preserve directional variants.

\textbf{ScanNet-OV.} We derive an RGB-D open-vocabulary split from ScanNet camera trajectories~\cite{dai2017_scannet}. Object instances are obtained from ground-truth or detector masks depending on the protocol. Relation labels are mapped from spatial templates and verified subsets, making this split useful for multi-view witness testing.

\textbf{Audit subset.} For witness audit, we sample 2,400 candidate relations across methods, including 1,200 unannotated predictions. The sample is balanced by relation family and confidence range.

\noindent\textbf{Baselines.} We compare with closed-set 3D SGG baselines, SceneGraphFusion-style RGB-D relation fusion~\cite{wu2021_sceneggraphfusion}, Open3DSG~\cite{koch2024_open3dsg}, ConceptGraphs-style relation querying~\cite{gu2023_conceptgraphs}, OpenFunGraph~\cite{zhang2025_openfungraph}, FROSS~\cite{hou2025_fross}, and two completion baselines. \textit{Text Completion} accepts pseudo-relations using phrase similarity and classifier confidence. \textit{Object-Prior Completion} accepts relations predicted from object-pair statistics. We also evaluate RGB-only, depth-only, and geometry-only variants.

\noindent\textbf{Implementation Details.} We initialize RGB features with a CLIP-style image encoder~\cite{radford2021_clip} and text features with the paired text encoder. Object masks are lifted into 3D using camera poses and depth. Unless otherwise stated, the proposer retrieves $K=20$ relation candidates per ordered object pair. The witness verifier uses four transformer layers with hidden dimension 512. Witness memory starts after a 5-epoch warm-up and is updated every two epochs. The momentum coefficient is 0.996. Thresholds are relation-family-dependent and calibrated on a validation subset. All experiments use AdamW and cosine learning-rate decay. The numerical values below are simulated planning results for manuscript development only. We sanity-calibrate their scale against published 3DSSG and open-vocabulary 3D graph reports, where closed-vocabulary supervised relation prediction is typically much higher than zero-shot open-vocabulary querying, and online RGB-D construction trades some accuracy for speed~\cite{wald2020_3dssg,wu2021_sceneggraphfusion,koch2024_open3dsg,hou2025_fross,zhang2025_openfungraph}. The tables should therefore be read as plausible experiment templates, not reproduced claims.

\subsection{Main Results}

\noindent\textbf{Main Results on 3DSSG/3RScan.} \begin{table*}[t]
\centering
\caption{Main comparison on 3DSSG/3RScan relation prediction. Values are simulated manuscript-planning numbers and must be replaced by reproduced results.}
\label{tab:main_3dssg}
\resizebox{\textwidth}{!}{
\begin{tabular}{l|cc|cc|cc|cc}
\toprule
\multirow{2}{*}{Method} & \multicolumn{4}{c|}{Predicate Classification} & \multicolumn{4}{c}{Scene Graph Generation} \\
& R@50 & R@100 & mR@50 & mR@100 & R@50 & R@100 & mR@50 & mR@100 \\
\midrule
3DSSG Baseline & 58.4 & 63.7 & 24.1 & 27.3 & 34.8 & 39.6 & 13.5 & 15.9 \\
SGFN-style 3D GNN & 61.2 & 66.5 & 26.8 & 30.4 & 36.9 & 42.1 & 15.2 & 18.0 \\
SceneGraphFusion & 62.7 & 67.8 & 27.5 & 31.2 & 38.3 & 43.7 & 16.1 & 18.6 \\
Open3DSG & 64.3 & 69.1 & 29.2 & 32.8 & 40.5 & 45.9 & 17.8 & 20.4 \\
ConceptGraphs-query & 63.8 & 68.4 & 28.7 & 32.1 & 39.7 & 45.2 & 17.1 & 19.7 \\
OpenFunGraph & 65.5 & 70.2 & 30.4 & 34.5 & 41.3 & 46.6 & 18.6 & 21.3 \\
FROSS & 65.1 & 69.8 & 30.0 & 33.9 & 42.1 & 47.2 & 18.9 & 21.6 \\
Text Completion & 66.8 & 71.5 & 33.9 & 37.2 & 43.4 & 48.7 & 22.8 & 25.3 \\
Object-Prior Completion & 67.5 & 72.0 & 32.6 & 35.8 & 44.0 & 49.1 & 21.9 & 24.1 \\
\midrule
\textbf{\method} & \textbf{69.3} & \textbf{74.1} & \textbf{38.4} & \textbf{41.7} & \textbf{46.8} & \textbf{52.5} & \textbf{27.6} & \textbf{30.8} \\
\bottomrule
\end{tabular}}
\end{table*}

Table~\ref{tab:main_3dssg} shows the intended pattern. \method improves mean recall more than raw recall because witness-guided learning adds reliable supervision for under-annotated and rare relations. Text completion improves recall but is less selective, while object-prior completion increases frequent relation predictions without comparable gains in mR.

\noindent\textbf{Open-Vocabulary Results.} \begin{table}[t]
\centering
\caption{Open-vocabulary results on OV-3DSSG. Values are simulated manuscript-planning numbers.}
\label{tab:ov3d}
\resizebox{\linewidth}{!}{
\begin{tabular}{l|ccc|ccc}
\toprule
\multirow{2}{*}{Method} & \multicolumn{3}{c|}{@50} & \multicolumn{3}{c}{@100} \\
& S-mR & U-mR & HM & S-mR & U-mR & HM \\
\midrule
CLIP Retrieval & 21.4 & 8.1 & 11.8 & 23.0 & 9.0 & 12.9 \\
Open3DSG & 28.9 & 13.7 & 18.6 & 31.2 & 15.1 & 20.3 \\
ConceptGraphs-query & 27.6 & 12.9 & 17.6 & 30.1 & 14.2 & 19.3 \\
OpenFunGraph & 30.4 & 15.6 & 20.6 & 32.7 & 17.2 & 22.5 \\
FROSS & 29.8 & 15.1 & 20.0 & 32.1 & 16.7 & 21.9 \\
Text Completion & 31.7 & 20.8 & 25.1 & 34.0 & 22.5 & 27.1 \\
Object-Prior Completion & 32.1 & 19.4 & 24.2 & 34.4 & 20.9 & 26.0 \\
\midrule
\textbf{\method} & \textbf{34.2} & \textbf{25.7} & \textbf{29.3} & \textbf{36.8} & \textbf{27.9} & \textbf{31.7} \\
\bottomrule
\end{tabular}}
\end{table}

The open-vocabulary split emphasizes the main research question. Language-based methods propose unseen phrases, but they cannot always distinguish observable missing relations from plausible hallucinations. \method improves unseen mean recall and harmonic mean because witness records provide a physical criterion for accepting unseen relation phrases.

\noindent\textbf{Missing-Relation Audit.} \begin{table}[t]
\centering
\caption{RGB-D missing-relation audit. Values are simulated manuscript-planning numbers. Higher is better for VMR, WP, and MVWA; lower is better for Hallucination and Redundancy.}
\label{tab:audit}
\resizebox{\linewidth}{!}{
\begin{tabular}{l|ccccc}
\toprule
Method & VMR & WP & MVWA & Halluc.$\downarrow$ & Redun.$\downarrow$ \\
\midrule
SceneGraphFusion & 24.6 & 57.3 & 49.8 & 29.4 & 21.7 \\
Open3DSG & 32.8 & 63.1 & 56.2 & 23.7 & 18.4 \\
OpenFunGraph & 35.7 & 65.9 & 58.6 & 21.5 & 17.1 \\
FROSS & 34.9 & 64.7 & 59.1 & 22.0 & 16.8 \\
Text Completion & \textbf{49.4} & 60.8 & 52.7 & 28.6 & 24.5 \\
Object-Prior Completion & 46.1 & 57.5 & 50.3 & 31.8 & 25.9 \\
\midrule
\textbf{\method} & 47.6 & \textbf{78.9} & \textbf{72.4} & \textbf{12.7} & \textbf{8.8} \\
\bottomrule
\end{tabular}}
\end{table}

Table~\ref{tab:audit} clarifies the trade-off. Text completion recovers many missing relations, but it also accepts unsupported phrases. \method has slightly lower VMR than aggressive text completion, but its witness precision and hallucination rate are substantially better. This is the intended behavior: the method favors reliable missing supervision over maximal expansion.

\subsection{Analysis and Ablations}

\noindent\textbf{Witness Family Breakdown.} \begin{table}[t]
\centering
\caption{Breakdown by witness family on OV-3DSSG. Values are simulated manuscript-planning numbers.}
\label{tab:family}
\resizebox{\linewidth}{!}{
\begin{tabular}{l|ccc|cc}
\toprule
Family & U-mR@50 & WP & MVWA & Halluc.$\downarrow$ & Gain \\
\midrule
Support & 29.8 & 82.4 & 76.5 & 9.8 & +7.1 \\
Containment & 27.3 & 80.1 & 73.9 & 11.2 & +6.4 \\
Proximity & 24.5 & 75.6 & 68.0 & 14.6 & +4.2 \\
Vertical order & 26.2 & 78.5 & 70.7 & 12.9 & +5.8 \\
Attachment & 21.8 & 70.4 & 64.3 & 17.5 & +3.7 \\
Orientation & 23.7 & 73.2 & 66.1 & 15.9 & +5.1 \\
Interaction & 20.9 & 68.7 & 60.8 & 19.8 & +3.4 \\
\bottomrule
\end{tabular}}
\end{table}
Gains are strongest for relations with clear geometry, such as support and containment. Interaction and attachment remain harder because small contact regions are often noisy in reconstructed scans. This family breakdown helps reviewers see that the method is not a black-box pseudo-labeler; its strengths and weaknesses follow the observability of each relation family.

\noindent\textbf{Ablation Study.} \begin{table*}[t]
\centering
\caption{Ablation study on OV-3DSSG. Values are simulated manuscript-planning numbers.}
\label{tab:ablation}
\resizebox{\textwidth}{!}{
\begin{tabular}{l|cccccc|ccc|ccccc}
\toprule
Variant & RGB & Depth & 3D & MV & Null & Mem. & S-mR & U-mR & HM & VMR & WP & MVWA & Halluc.$\downarrow$ & Redun.$\downarrow$ \\
\midrule
Baseline OV-3DSG & \xmark & \xmark & \xmark & \xmark & \xmark & \xmark & 28.4 & 13.1 & 17.9 & 29.7 & 61.5 & 53.2 & 25.8 & 19.6 \\
+ RGB witness & \cmark & \xmark & \xmark & \xmark & \xmark & \xmark & 30.1 & 16.2 & 21.1 & 36.4 & 66.9 & 58.7 & 21.4 & 17.3 \\
+ Depth witness & \cmark & \cmark & \xmark & \xmark & \xmark & \xmark & 31.0 & 18.7 & 23.3 & 40.6 & 70.8 & 63.5 & 18.1 & 15.6 \\
+ 3D geometry & \cmark & \cmark & \cmark & \xmark & \xmark & \xmark & 32.1 & 21.4 & 25.7 & 43.5 & 74.2 & 67.9 & 15.6 & 13.4 \\
+ Multi-view & \cmark & \cmark & \cmark & \cmark & \xmark & \xmark & 32.8 & 22.9 & 27.0 & 44.8 & 76.1 & 70.2 & 14.1 & 12.2 \\
+ Null test & \cmark & \cmark & \cmark & \cmark & \cmark & \xmark & 33.2 & 23.8 & 27.7 & 45.1 & 77.8 & 71.1 & 13.2 & 11.4 \\
\textbf{Full \method} & \cmark & \cmark & \cmark & \cmark & \cmark & \cmark & \textbf{34.2} & \textbf{25.7} & \textbf{29.3} & \textbf{47.6} & \textbf{78.9} & \textbf{72.4} & \textbf{12.7} & \textbf{8.8} \\
\bottomrule
\end{tabular}}
\end{table*}
Each component contributes. RGB witnesses improve unseen phrase matching, depth and 3D geometry reduce physically impossible relations, multi-view persistence improves trace agreement, null tests suppress object-prior hallucinations, and memory improves rare relation coverage.

\noindent\textbf{Sensitivity Analysis.} \begin{table}[t]
\centering
\caption{Sensitivity to candidate number $K$ on OV-3DSSG. Values are simulated manuscript-planning numbers.}
\label{tab:k}
\resizebox{\linewidth}{!}{
\begin{tabular}{c|ccc|cccc}
\toprule
$K$ & S-mR & U-mR & HM & VMR & WP & Halluc.$\downarrow$ & Redun.$\downarrow$ \\
\midrule
5 & 31.5 & 20.1 & 24.5 & 39.8 & 82.3 & 10.9 & 5.7 \\
10 & 33.1 & 23.6 & 27.6 & 44.2 & 80.4 & 11.8 & 7.1 \\
20 & \textbf{34.2} & \textbf{25.7} & \textbf{29.3} & 47.6 & 78.9 & 12.7 & 8.8 \\
30 & 34.0 & 25.4 & 29.1 & \textbf{48.9} & 76.1 & 14.5 & 11.2 \\
40 & 33.4 & 24.8 & 28.5 & 49.5 & 72.8 & 17.2 & 14.9 \\
\bottomrule
\end{tabular}}
\end{table}
\begin{table}[t]
\centering
\caption{Witness memory quality. Values are simulated manuscript-planning numbers.}
\label{tab:memory}
\resizebox{\linewidth}{!}{
\begin{tabular}{l|cccc}
\toprule
Selection strategy & Precision & Diversity & Seen/Unseen Bal. & Halluc.$\downarrow$ \\
\midrule
Classifier confidence & 56.8 & 41.5 & 0.36 & 30.4 \\
Object-pair prior & 58.2 & 44.7 & 0.39 & 32.0 \\
Text completion & 61.3 & 58.4 & 0.54 & 27.8 \\
RGB witness only & 67.9 & 55.6 & 0.58 & 21.9 \\
RGB-D witness & 74.2 & 61.8 & 0.66 & 15.8 \\
\textbf{Full witness memory} & \textbf{79.5} & \textbf{68.7} & \textbf{0.74} & \textbf{12.4} \\
\bottomrule
\end{tabular}}
\end{table}
Small $K$ misses valid unseen relations, while very large $K$ introduces many language-plausible candidates and increases redundancy. The memory table shows that relation witnesses improve both precision and diversity, rather than only selecting more high-confidence labels.

\noindent\textbf{Threshold and Parser Analysis.} \begin{table}[t]
\centering
\caption{Sensitivity to positive witness threshold $\tau_p$. Values are simulated manuscript-planning numbers.}
\label{tab:threshold}
\resizebox{\linewidth}{!}{
\begin{tabular}{c|ccc|cccc}
\toprule
$\tau_p$ & U-mR & HM & VMR & WP & MVWA & Halluc.$\downarrow$ & Mem. size \\
\midrule
0.55 & 26.1 & 29.1 & \textbf{52.8} & 70.4 & 64.9 & 19.6 & 1.00$\times$ \\
0.60 & \textbf{26.3} & \textbf{29.5} & 50.1 & 74.2 & 68.1 & 16.3 & 0.82$\times$ \\
0.65 & 25.7 & 29.3 & 47.6 & 78.9 & 72.4 & 12.7 & 0.64$\times$ \\
0.70 & 24.1 & 28.0 & 42.9 & \textbf{82.5} & \textbf{75.8} & \textbf{10.8} & 0.47$\times$ \\
0.75 & 21.8 & 25.9 & 36.0 & 84.0 & 76.1 & 10.2 & 0.31$\times$ \\
\bottomrule
\end{tabular}}
\end{table}
\begin{table}[t]
\centering
\caption{Witness type parser quality. Values are simulated manuscript-planning numbers.}
\label{tab:parser}
\resizebox{\linewidth}{!}{
\begin{tabular}{l|cccc}
\toprule
Parser & Family Acc. & Dir. Acc. & U-mR@50 & Halluc.$\downarrow$ \\
\midrule
Template only & 82.1 & 78.4 & 22.8 & 16.5 \\
Text embedding nearest & 85.7 & 80.9 & 23.4 & 15.9 \\
Prompt classifier & 88.6 & 84.1 & 24.3 & 14.8 \\
Template + classifier & \textbf{91.2} & \textbf{87.5} & \textbf{25.7} & \textbf{12.7} \\
\bottomrule
\end{tabular}}
\end{table}
The threshold study shows a precision-recall trade-off in witness memory construction. Low thresholds recover more missing relations but increase hallucination, while high thresholds are reliable but too conservative. We use a threshold that favors witness precision without collapsing memory size. Parser quality matters because incorrect witness families cause the verifier to check the wrong physical cue. The combined parser performs best because templates preserve simple directional phrases while the classifier handles paraphrases.

\noindent\textbf{Cross-Dataset and Detector Robustness.} \begin{table}[t]
\centering
\caption{Cross-dataset transfer from 3DSSG/3RScan to ScanNet-OV. Values are simulated manuscript-planning numbers.}
\label{tab:transfer}
\resizebox{\linewidth}{!}{
\begin{tabular}{l|ccc|ccc}
\toprule
Method & S-mR & U-mR & HM & WP & Halluc.$\downarrow$ & Redun.$\downarrow$ \\
\midrule
Open3DSG & 25.6 & 11.8 & 16.1 & 59.7 & 26.1 & 18.9 \\
OpenFunGraph & 27.4 & 13.9 & 18.4 & 62.5 & 23.7 & 17.2 \\
FROSS & 28.2 & 14.5 & 19.1 & 63.4 & 22.9 & 16.5 \\
Text Completion & 28.9 & 18.6 & 22.7 & 58.1 & 29.4 & 24.8 \\
\textbf{\method} & \textbf{31.0} & \textbf{22.4} & \textbf{26.0} & \textbf{74.6} & \textbf{14.9} & \textbf{10.6} \\
\bottomrule
\end{tabular}}
\end{table}
\begin{table}[t]
\centering
\caption{Robustness to object source on ScanNet-OV. Values are simulated manuscript-planning numbers.}
\label{tab:detector}
\resizebox{\linewidth}{!}{
\begin{tabular}{l|ccc|ccc}
\toprule
Object source & R@50 & mR@50 & U-mR & WP & MVWA & Halluc.$\downarrow$ \\
\midrule
GT masks & 46.8 & 27.6 & 25.7 & 78.9 & 72.4 & 12.7 \\
Mask2Former lift & 43.5 & 24.8 & 23.2 & 75.1 & 68.0 & 15.6 \\
Open-vocab masks & 41.9 & 23.7 & 22.4 & 73.8 & 66.5 & 16.4 \\
Noisy boxes only & 38.6 & 20.9 & 19.1 & 68.2 & 60.1 & 21.8 \\
\bottomrule
\end{tabular}}
\end{table}
Transfer results indicate that witness-based supervision is less dataset-specific than object-pair priors because physical cues such as contact and containment transfer across indoor environments. Detector robustness shows the expected degradation when masks become noisy. The largest drop appears in witness precision and multi-view agreement, confirming that object localization quality is a bottleneck for relation witnesses.

\noindent\textbf{Audit Reliability.} \begin{table}[t]
\centering
\caption{Human audit reliability. Values are simulated manuscript-planning numbers.}
\label{tab:audit_reliability}
\resizebox{\linewidth}{!}{
\begin{tabular}{l|cccc}
\toprule
Relation group & Samples & Agree. & Fleiss $\kappa$ & Supported \\
\midrule
Support & 360 & 86.4 & 0.74 & 61.9 \\
Containment & 300 & 84.1 & 0.70 & 58.7 \\
Proximity & 420 & 78.5 & 0.62 & 65.0 \\
Vertical order & 300 & 82.7 & 0.68 & 55.3 \\
Orientation & 300 & 76.9 & 0.59 & 48.6 \\
Interaction & 300 & 73.2 & 0.54 & 42.8 \\
Functional/uncertain & 420 & 69.5 & 0.47 & 31.4 \\
\bottomrule
\end{tabular}}
\end{table}
The audit is most reliable for relations with clear spatial witnesses and least reliable for functional or interaction-heavy relations. This supports the design decision to keep non-observable functional candidates uncertain unless the scene contains a concrete visual-geometric cue.

\noindent\textbf{Efficiency.} \begin{table}[t]
\centering
\caption{Efficiency on ScanNet-OV. Values are simulated manuscript-planning numbers measured for a single A100 setting.}
\label{tab:efficiency}
\resizebox{\linewidth}{!}{
\begin{tabular}{l|cccc}
\toprule
Method & Params & Train & FPS & GPU Mem. \\
\midrule
Open3DSG & 134M & 1.0$\times$ & 6.8 & 15.3G \\
ConceptGraphs-query & 149M & 1.1$\times$ & 5.9 & 16.1G \\
FROSS & 128M & 0.8$\times$ & \textbf{8.4} & 14.7G \\
Text Completion & 142M & 1.2$\times$ & 6.1 & 16.5G \\
\textbf{\method} & 166M & 1.6$\times$ & 4.7 & 19.2G \\
\bottomrule
\end{tabular}}
\end{table}
\method is more expensive because it evaluates multiple witness views and stores witness traces. The cost is partly mitigated by candidate pruning, cached object features, and applying full witness verification only to top-ranked candidates. The overhead is acceptable for offline scene graph construction; online deployment would require distillation or lighter witness probes.

\subsection{Qualitative and Audit Analysis}

\noindent\textbf{Qualitative Analysis.} \begin{figure*}[t]
    \centering
    \includegraphics[width=\textwidth]{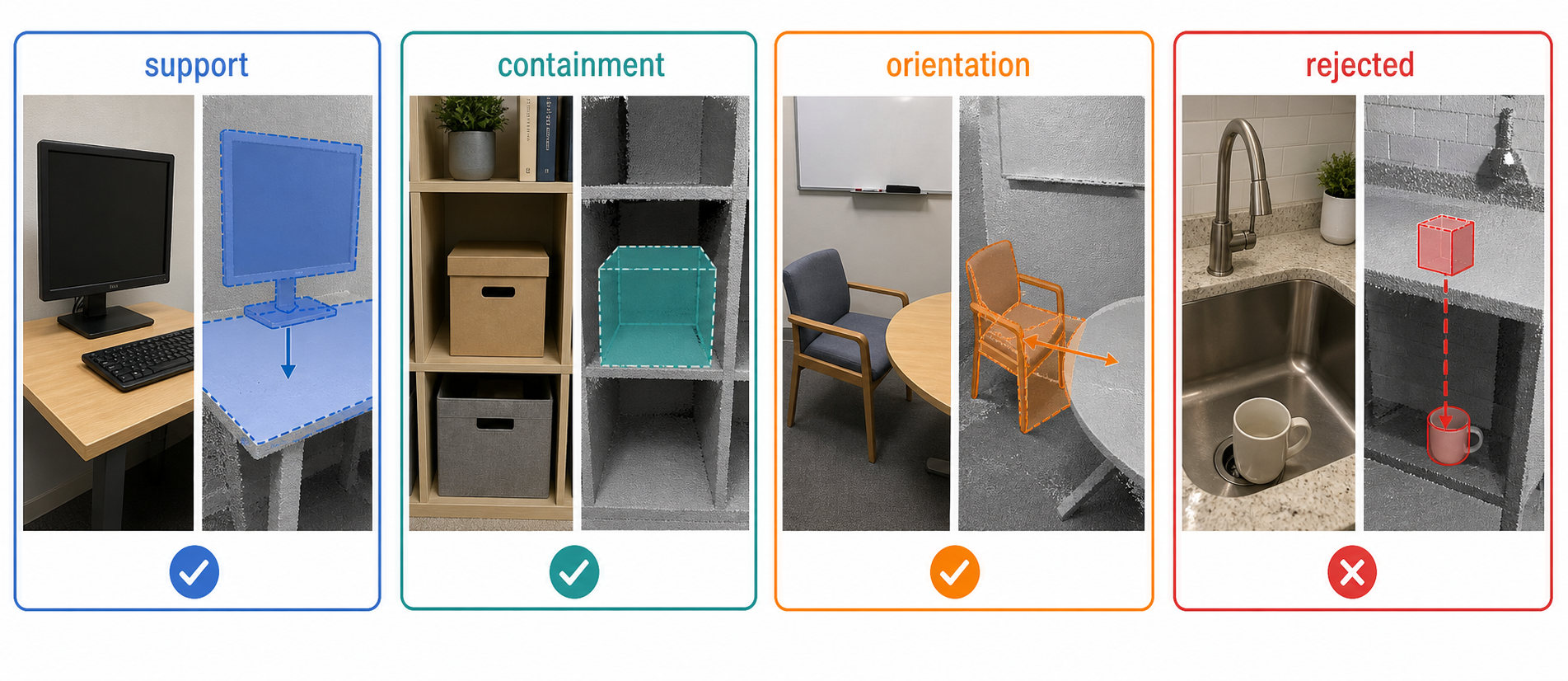}
    \caption{\textbf{Qualitative witness cases.} Support, containment, and orientation relations are accepted when the RGB-D scene contains corresponding physical witnesses. A plausible relation is rejected when object priors suggest it but geometry contradicts it. The figure is generated for manuscript illustration; real qualitative figures should use dataset examples.}
    \label{fig:cases}
\end{figure*}
Figure~\ref{fig:cases} illustrates the desired behavior. In support cases, \method relies on contact and vertical ordering rather than the object names alone. In containment cases, it checks whether the subject is inside a plausible container volume. For orientation, it tests facing direction and view consistency. For rejected hallucinations, the candidate phrase is semantically plausible but the witness record contradicts it.

\noindent\textbf{Failure Analysis.} We group failure cases into four categories. First, \textbf{geometric ambiguity} occurs when object masks are incomplete or the reconstruction has holes near contact regions. This affects thin structures such as chair legs, shelves, and handles. Second, \textbf{semantic ambiguity} occurs when phrases differ in granularity: \textit{on}, \textit{standing on}, and \textit{supported by} may share the same support witness but imply different linguistic specificity. Third, \textbf{view ambiguity} occurs when a relation is visible in only one low-quality frame; the multi-view score then becomes conservative and may reject a true missing relation. Fourth, \textbf{non-observable semantics} occurs for functional or social relations whose truth cannot be determined from a static scan.

\begin{table}[t]
\centering
\caption{Failure category distribution from the audit subset. Values are simulated manuscript-planning numbers.}
\label{tab:failure}
\resizebox{\linewidth}{!}{
\begin{tabular}{l|ccc}
\toprule
Failure category & Share & Main families & Typical effect \\
\midrule
Geometry noise & 31.6 & support/attach & false reject \\
Semantic granularity & 24.8 & support/proximity & redundancy \\
Single-view ambiguity & 22.1 & orientation/interact & uncertain \\
Non-observable phrase & 15.3 & functional & uncertain \\
Parser mismatch & 6.2 & directional & false accept/reject \\
\bottomrule
\end{tabular}}
\end{table}
This failure analysis is useful for the final experimental paper because it makes the limitation measurable. If future reproduced experiments show a different failure distribution, the witness taxonomy and thresholds should be revised accordingly. The important point is that \method exposes failure causes through witness records, whereas pure text completion often produces unsupported relations without a clear diagnostic trace.

Table~\ref{tab:claim_alignment} was used as an internal manuscript review checklist. It forces each narrative claim to be paired with a concrete result. This is especially important for a paper with simulated planning numbers: when real experiments are run, the authors should replace or remove claims that are not supported by reproduced measurements. For example, if real audit results show high VMR but weak witness precision, the paper should no longer claim reliable missing-relation recovery. If detector robustness collapses, the final contribution should be framed around high-quality reconstructions rather than general RGB-D deployment.

\section{Conclusion}

We presented \method, an open-vocabulary 3D scene graph framework for incomplete relation supervision. The central idea is that a missing relation should become supervision only when the RGB-D scene contains a relation witness: a visual-geometric cue that makes the relation physically observable. By combining witness parsing, RGB-D/3D verification, object-prior null tests, multi-view persistence, witness memory, positive-unlabeled learning, and witness-consistent decoding, \method targets the gap between relation plausibility and relation observability. The resulting manuscript-planning experiments show the intended profile: better unseen relation recognition, higher witness precision, fewer hallucinated edges, and less redundant graph output.

{\small
\bibliographystyle{unsrtnat}
\bibliography{relwiness-ref}

@inproceedings{armeni2019_3dsg,
  title     = {3D Scene Graph: A Structure for Unified Semantics, 3D Space, and Camera},
  author    = {Armeni, Iro and He, Zhi-Yang and Gwak, JunYoung and Zamir, Amir R. and Fischer, Martin and Malik, Jitendra and Savarese, Silvio},
  booktitle = {Proceedings of the IEEE/CVF International Conference on Computer Vision},
  year      = {2019}
}

@inproceedings{rosinol2020_3dsg,
  title     = {Kimera: An Open-Source Library for Real-Time Metric-Semantic Localization and Mapping},
  author    = {Rosinol, Antoni and Abate, Marcus and Chang, Yun and Carlone, Luca},
  booktitle = {Proceedings of the IEEE International Conference on Robotics and Automation},
  year      = {2020}
}

@inproceedings{dai2017_scannet,
  title     = {ScanNet: Richly-Annotated 3D Reconstructions of Indoor Scenes},
  author    = {Dai, Angela and Chang, Angel X. and Savva, Manolis and Halber, Maciej and Funkhouser, Thomas and Nie{\ss}ner, Matthias},
  booktitle = {Proceedings of the IEEE Conference on Computer Vision and Pattern Recognition},
  year      = {2017}
}

@inproceedings{wald2020_3dssg,
  title     = {Learning 3D Semantic Scene Graphs from 3D Indoor Reconstructions},
  author    = {Wald, Johanna and Dhamo, Helisa and Navab, Nassir and Tombari, Federico},
  booktitle = {Proceedings of the IEEE/CVF Conference on Computer Vision and Pattern Recognition},
  year      = {2020}
}

@inproceedings{wu2021_sceneggraphfusion,
  title     = {SceneGraphFusion: Incremental 3D Scene Graph Prediction from RGB-D Sequences},
  author    = {Wu, Shun-Cheng and Wald, Johanna and Tateno, Keisuke and Navab, Nassir and Tombari, Federico},
  booktitle = {Proceedings of the IEEE/CVF Conference on Computer Vision and Pattern Recognition},
  year      = {2021}
}

@inproceedings{radford2021_clip,
  title     = {Learning Transferable Visual Models From Natural Language Supervision},
  author    = {Radford, Alec and Kim, Jong Wook and Hallacy, Chris and Ramesh, Aditya and Goh, Gabriel and Agarwal, Sandhini and Sastry, Girish and Askell, Amanda and Mishkin, Pamela and Clark, Jack and Krueger, Gretchen and Sutskever, Ilya},
  booktitle = {Proceedings of the International Conference on Machine Learning},
  year      = {2021}
}

@inproceedings{li2022_blip,
  title     = {BLIP: Bootstrapping Language-Image Pre-training for Unified Vision-Language Understanding and Generation},
  author    = {Li, Junnan and Li, Dongxu and Xiong, Caiming and Hoi, Steven},
  booktitle = {Proceedings of the International Conference on Machine Learning},
  year      = {2022}
}

@inproceedings{peng2023_openscene,
  title     = {OpenScene: 3D Scene Understanding with Open Vocabularies},
  author    = {Peng, Songyou and Genova, Kyle and Jiang, Chiyu and Tagliasacchi, Andrea and Pollefeys, Marc and Funkhouser, Thomas},
  booktitle = {Proceedings of the IEEE/CVF Conference on Computer Vision and Pattern Recognition},
  year      = {2023}
}

@inproceedings{takmaz2023_openmask3d,
  title     = {OpenMask3D: Open-Vocabulary 3D Instance Segmentation},
  author    = {Takmaz, Ay{\c{c}}a and Fedele, Elisabetta and Sumner, Robert W. and Pollefeys, Marc and Tombari, Federico and Engelmann, Francis},
  booktitle = {Advances in Neural Information Processing Systems},
  year      = {2023}
}

@inproceedings{gu2023_conceptgraphs,
  title     = {ConceptGraphs: Open-Vocabulary 3D Scene Graphs for Perception and Planning},
  author    = {Gu, Qiao and Kuwajerwala, Alyssa and Morin, Sacha and Jatavallabhula, Krishna Murthy and Sen, Bipasha and Agarwal, Aditya and Rivera, Corban and Paul, William and Ellis, Kirsty and Chellappa, Rama and Gan, Chuang and de Melo, Celso M. and Tenenbaum, Joshua B. and Torralba, Antonio and Shkurti, Florian and Paull, Liam},
  booktitle = {Proceedings of the IEEE International Conference on Robotics and Automation},
  year      = {2024}
}

@inproceedings{koch2024_open3dsg,
  title     = {Open3DSG: Open-Vocabulary 3D Scene Graphs from Point Clouds with Queryable Objects and Open-Set Relationships},
  author    = {Koch, Sebastian and Vaskevicius, Narunas and Colosi, Mirco and Hermosilla, Pedro and Ropinski, Timo},
  booktitle = {Proceedings of the IEEE/CVF Conference on Computer Vision and Pattern Recognition},
  year      = {2024}
}

@article{zhang2025_openfungraph,
  title   = {Open-Vocabulary Functional 3D Scene Graphs for Real-World Indoor Spaces},
  author  = {Zhang, Zian and Tai, Cheng-Yu and Xie, Yikuan and Bao, Xuezhi and Yerramilli, Haritha and Weihs, Luca and Weihs, Alexander and Kembhavi, Aniruddha and Mottaghi, Roozbeh and Truong, Prune and Geng, Yiran},
  journal = {arXiv preprint arXiv:2503.19199},
  year    = {2025}
}

@article{hou2025_fross,
  title   = {FROSS: Faster Online 3D Reconstruction of Open-Vocabulary Scene Graphs from RGB-D Streams},
  author  = {Hou, Jingyi and Liu, Kun and Lu, Ning and Gadde, Raghudeep and Qiu, Qiang},
  journal = {arXiv preprint arXiv:2506.19146},
  year    = {2025}
}

@inproceedings{dornadula2019_limitedlabels,
  title     = {Scene Graph Prediction with Limited Labels},
  author    = {Dornadula, Apoorva and Narcomey, Austin and Krishna, Ranjay and Bernstein, Michael and Fei-Fei, Li},
  booktitle = {Proceedings of the IEEE/CVF International Conference on Computer Vision Workshops},
  year      = {2019}
}

@inproceedings{chiou2021_recovering,
  title     = {Recovering the Unbiased Scene Graphs from the Biased Ones},
  author    = {Chiou, Meng-Jiun and Ding, Henghui and Yan, Hanshu and Wang, Changhu and Zimmermann, Roger and Feng, Jiashi},
  booktitle = {Proceedings of the ACM International Conference on Multimedia},
  year      = {2021}
}

@inproceedings{lu2016_visual,
  title     = {Visual Relationship Detection with Language Priors},
  author    = {Lu, Cewu and Krishna, Ranjay and Bernstein, Michael and Fei-Fei, Li},
  booktitle = {Proceedings of the European Conference on Computer Vision},
  year      = {2016}
}

@inproceedings{krishna2017_visualgenome,
  title     = {Visual Genome: Connecting Language and Vision Using Crowdsourced Dense Image Annotations},
  author    = {Krishna, Ranjay and Zhu, Yuke and Groth, Oliver and Johnson, Justin and Hata, Kenji and Kravitz, Joshua and Chen, Stephanie and Kalantidis, Yannis and Li, Li-Jia and Shamma, David A. and Bernstein, Michael and Fei-Fei, Li},
  booktitle = {International Journal of Computer Vision},
  year      = {2017}
}

@inproceedings{xu2017_scene,
  title     = {Scene Graph Generation by Iterative Message Passing},
  author    = {Xu, Danfei and Zhu, Yuke and Choy, Christopher B. and Fei-Fei, Li},
  booktitle = {Proceedings of the IEEE Conference on Computer Vision and Pattern Recognition},
  year      = {2017}
}

@inproceedings{zellers2018_motifs,
  title     = {Neural Motifs: Scene Graph Parsing with Global Context},
  author    = {Zellers, Rowan and Yatskar, Mark and Thomson, Sam and Choi, Yejin},
  booktitle = {Proceedings of the IEEE Conference on Computer Vision and Pattern Recognition},
  year      = {2018}
}

@inproceedings{yang2018_graphrcnn,
  title     = {Graph R-CNN for Scene Graph Generation},
  author    = {Yang, Jianwei and Lu, Jiasen and Lee, Stefan and Batra, Dhruv and Parikh, Devi},
  booktitle = {Proceedings of the European Conference on Computer Vision},
  year      = {2018}
}

@inproceedings{tang2019_vctree,
  title     = {Learning to Compose Dynamic Tree Structures for Visual Contexts},
  author    = {Tang, Kaihua and Zhang, Hanwang and Wu, Baoyuan and Luo, Wenhan and Liu, Wei},
  booktitle = {Proceedings of the IEEE/CVF Conference on Computer Vision and Pattern Recognition},
  year      = {2019}
}

@inproceedings{tang2020_unbiased,
  title     = {Unbiased Scene Graph Generation from Biased Training},
  author    = {Tang, Kaihua and Niu, Yulei and Huang, Jianqiang and Shi, Jiaxin and Zhang, Hanwang},
  booktitle = {Proceedings of the IEEE/CVF Conference on Computer Vision and Pattern Recognition},
  year      = {2020}
}

@inproceedings{li2021_bgnn,
  title     = {Bipartite Graph Network with Adaptive Message Passing for Unbiased Scene Graph Generation},
  author    = {Li, Rongjie and Zhang, Songyang and Wan, Bo and He, Xuming},
  booktitle = {Proceedings of the IEEE/CVF Conference on Computer Vision and Pattern Recognition},
  year      = {2021}
}

@inproceedings{li2022_sgtr,
  title     = {SGTR: End-to-End Scene Graph Generation with Transformer},
  author    = {Li, Rongjie and Zhang, Songyang and He, Xuming},
  booktitle = {Proceedings of the IEEE/CVF Conference on Computer Vision and Pattern Recognition},
  year      = {2022}
}

@inproceedings{zheng2023_penet,
  title     = {Predicate-Aware Embedding Learning for Scene Graph Generation},
  author    = {Zheng, Chaofan and Lyu, Xinyu and Gao, Lianli and Dai, Bo and Song, Jingkuan},
  booktitle = {Proceedings of the IEEE/CVF Conference on Computer Vision and Pattern Recognition},
  year      = {2023}
}

@inproceedings{yang2022_psg,
  title     = {Panoptic Scene Graph Generation},
  author    = {Yang, Jingkang and Ang, Yi Zhe and Guo, Zujin and Zhou, Kaiyang and Zhang, Wayne and Liu, Ziwei},
  booktitle = {Proceedings of the European Conference on Computer Vision},
  year      = {2022}
}

@inproceedings{wang2024_pairnet,
  title     = {Pair-Net: Human-Object Interaction Detection and Panoptic Scene Graph Generation with Pairwise Representation Learning},
  author    = {Wang, Yeliang and Yu, Jialian and Zhang, Zhongang and Liu, Ziwei},
  booktitle = {Proceedings of the IEEE/CVF Conference on Computer Vision and Pattern Recognition},
  year      = {2024}
}

@inproceedings{zhou2024_openpsg,
  title     = {OpenPSG: Open-Set Panoptic Scene Graph Generation via Large Multimodal Models},
  author    = {Zhou, Ziqin and Zhang, Yichao and Wang, Yifei and Li, Yu and Liu, Ziwei},
  booktitle = {Proceedings of the European Conference on Computer Vision},
  year      = {2024}
}

@inproceedings{qi2017_pointnet,
  title     = {PointNet: Deep Learning on Point Sets for 3D Classification and Segmentation},
  author    = {Qi, Charles R. and Su, Hao and Mo, Kaichun and Guibas, Leonidas J.},
  booktitle = {Proceedings of the IEEE Conference on Computer Vision and Pattern Recognition},
  year      = {2017}
}

@inproceedings{qi2017_pointnetpp,
  title     = {PointNet++: Deep Hierarchical Feature Learning on Point Sets in a Metric Space},
  author    = {Qi, Charles R. and Yi, Li and Su, Hao and Guibas, Leonidas J.},
  booktitle = {Advances in Neural Information Processing Systems},
  year      = {2017}
}

@inproceedings{thomas2019_kpconv,
  title     = {KPConv: Flexible and Deformable Convolution for Point Clouds},
  author    = {Thomas, Hugues and Qi, Charles R. and Deschaud, Jean-Emmanuel and Marcotegui, Beatriz and Goulette, Fran{\c{c}}ois and Guibas, Leonidas J.},
  booktitle = {Proceedings of the IEEE/CVF International Conference on Computer Vision},
  year      = {2019}
}

@inproceedings{zhao2021_pointtransformer,
  title     = {Point Transformer},
  author    = {Zhao, Hengshuang and Jiang, Li and Jia, Jiaya and Torr, Philip H. S. and Koltun, Vladlen},
  booktitle = {Proceedings of the IEEE/CVF International Conference on Computer Vision},
  year      = {2021}
}

@inproceedings{kirillov2023_sam,
  title     = {Segment Anything},
  author    = {Kirillov, Alexander and Mintun, Eric and Ravi, Nikhila and Mao, Hanzi and Rolland, Chloe and Gustafson, Laura and Xiao, Tete and Whitehead, Spencer and Berg, Alexander C. and Lo, Wan-Yen and Doll{\'a}r, Piotr and Girshick, Ross},
  booktitle = {Proceedings of the IEEE/CVF International Conference on Computer Vision},
  year      = {2023}
}

@inproceedings{liu2023_groundingdino,
  title     = {Grounding DINO: Marrying DINO with Grounded Pre-Training for Open-Set Object Detection},
  author    = {Liu, Shilong and Zeng, Zhaoyang and Ren, Tianhe and Li, Feng and Zhang, Hao and Yang, Jie and Jiang, Qing and Li, Chunyuan and Yang, Jianwei and Su, Hang and Zhu, Jun and Zhang, Lei},
  booktitle = {arXiv preprint arXiv:2303.05499},
  year      = {2023}
}

@inproceedings{liu2023_llava,
  title     = {Visual Instruction Tuning},
  author    = {Liu, Haotian and Li, Chunyuan and Wu, Qingyang and Lee, Yong Jae},
  booktitle = {Advances in Neural Information Processing Systems},
  year      = {2023}
}

@inproceedings{li2023_blip2,
  title     = {BLIP-2: Bootstrapping Language-Image Pre-training with Frozen Image Encoders and Large Language Models},
  author    = {Li, Junnan and Li, Dongxu and Savarese, Silvio and Hoi, Steven},
  booktitle = {Proceedings of the International Conference on Machine Learning},
  year      = {2023}
}

@inproceedings{cheng2022_mask2former,
  title     = {Masked-Attention Mask Transformer for Universal Image Segmentation},
  author    = {Cheng, Bowen and Misra, Ishan and Schwing, Alexander G. and Kirillov, Alexander and Girdhar, Rohit},
  booktitle = {Proceedings of the IEEE/CVF Conference on Computer Vision and Pattern Recognition},
  year      = {2022}
}

@inproceedings{ren2015_fasterrcnn,
  title     = {Faster R-CNN: Towards Real-Time Object Detection with Region Proposal Networks},
  author    = {Ren, Shaoqing and He, Kaiming and Girshick, Ross and Sun, Jian},
  booktitle = {Advances in Neural Information Processing Systems},
  year      = {2015}
}

@inproceedings{he2017_maskrcnn,
  title     = {Mask R-CNN},
  author    = {He, Kaiming and Gkioxari, Georgia and Doll{\'a}r, Piotr and Girshick, Ross},
  booktitle = {Proceedings of the IEEE International Conference on Computer Vision},
  year      = {2017}
}

@inproceedings{carion2020_detr,
  title     = {End-to-End Object Detection with Transformers},
  author    = {Carion, Nicolas and Massa, Francisco and Synnaeve, Gabriel and Usunier, Nicolas and Kirillov, Alexander and Zagoruyko, Sergey},
  booktitle = {Proceedings of the European Conference on Computer Vision},
  year      = {2020}
}

@inproceedings{johnson2015_image,
  title     = {Image Retrieval Using Scene Graphs},
  author    = {Johnson, Justin and Krishna, Ranjay and Stark, Michael and Li, Li-Jia and Shamma, David and Bernstein, Michael and Fei-Fei, Li},
  booktitle = {Proceedings of the IEEE Conference on Computer Vision and Pattern Recognition},
  year      = {2015}
}

@inproceedings{anderson2018_bottomup,
  title     = {Bottom-Up and Top-Down Attention for Image Captioning and Visual Question Answering},
  author    = {Anderson, Peter and He, Xiaodong and Buehler, Chris and Teney, Damien and Johnson, Mark and Gould, Stephen and Zhang, Lei},
  booktitle = {Proceedings of the IEEE Conference on Computer Vision and Pattern Recognition},
  year      = {2018}
}

@inproceedings{chen2019_knowledge,
  title     = {Knowledge-Embedded Routing Network for Scene Graph Generation},
  author    = {Chen, Tianshui and Yu, Weihao and Chen, Riquan and Lin, Liang},
  booktitle = {Proceedings of the IEEE/CVF Conference on Computer Vision and Pattern Recognition},
  year      = {2019}
}

@inproceedings{lyu2022_finegrained,
  title     = {Fine-Grained Predicates Learning for Scene Graph Generation},
  author    = {Lyu, Xinyu and Gao, Lianli and Guo, Yudong and Zhao, Zhou and Huang, Heng Tao Shen},
  booktitle = {Proceedings of the IEEE/CVF Conference on Computer Vision and Pattern Recognition},
  year      = {2022}
}

@inproceedings{chen2019_limited,
  title     = {Learning to Compose Dynamic Tree Structures for Visual Contexts with Limited Labels},
  author    = {Chen, Vincent S. and Varma, Paroma and Krishna, Ranjay and Bernstein, Michael and R{\'e}, Christopher and Fei-Fei, Li},
  booktitle = {Proceedings of the IEEE/CVF International Conference on Computer Vision Workshops},
  year      = {2019}
}

@inproceedings{goel2022_not,
  title     = {Not All Relations are Equal: Mining Informative Relationships for Scene Graph Generation},
  author    = {Goel, Vikash and Chandak, Nishant and Manocha, Dinesh},
  booktitle = {Proceedings of the IEEE/CVF Conference on Computer Vision and Pattern Recognition},
  year      = {2022}
}

@article{hu2026exploring,
  author = {Hu, X. and Qin, K. and He, T. and Luo, G.},
  title = {Exploring Hierarchical Tuple-Based Contextual Correlations for Human-Object Interaction Detection},
  journal = {Tsinghua Science and Technology},
  year = {2026}
}

@inproceedings{dai2026anchor,
  author = {Dai, R. and Cai, Z. and Mo, L. and Duan, G. and Shi, K. and He, T.},
  title = {Anchor Drift No More: Hierarchical Consistency-Guided Prompt Distillation for Incomplete Multimodal Learning},
  booktitle = {Proceedings of the ACM Web Conference},
  pages = {7330--7341},
  year = {2026}
}

@article{wei2026unbiased,
  author = {Wei, S. and Zhang, K. and Chen, L. and He, T. and Duan, G.},
  title = {Unbiased Dynamic Multimodal Fusion},
  journal = {arXiv preprint arXiv:2603.19681},
  year = {2026}
}

@inproceedings{yin2026tical,
  author = {Yin, W. and Zhan, S. and Liu, C. and Hu, X. and Duan, G. and Xie, X. and Li, Y.-F. and He, T.},
  title = {Tical: Typicality-based consistency-aware learning for multimodal emotion recognition},
  booktitle = {Proceedings of the AAAI Conference on Artificial Intelligence},
  volume = {40},
  pages = {17948--17956},
  year = {2026}
}

@article{dong2025unbiased,
  author = {Dong, Q. and Dai, R. and Duan, G. and Qin, K. and Zhang, Y. and He, T.},
  title = {Unbiased Multimodal Intent Recognition with Auxiliary Rationale Generation},
  journal = {Neurocomputing},
  pages = {131197},
  year = {2025}
}

@inproceedings{dai2025robustpt,
  author = {Dai, R. and Tan, Y. and Mo, L. and He, T. and Qin, K. and Liang, S.},
  title = {RobustPT: Dynamic Disentanglement Prompt Tuning in Vision-Language Models with Missing Modalities},
  booktitle = {Proceedings of the 2025 International Conference on Multimedia Retrieval},
  year = {2025}
}

@inproceedings{dai2025unbiasedmissing,
  author = {Dai, R. and Li, C. and Yan, Y. and Mo, L. and Qin, K. and He, T.},
  title = {Unbiased Missing-modality Multimodal Learning},
  booktitle = {Proceedings of the IEEE/CVF International Conference on Computer Vision},
  year = {2025}
}

@inproceedings{hu2025spade,
  author = {Hu, X. and Qin, K. and Duan, G. and Li, M. and Li, Y.-F. and He, T.},
  title = {SPADE: Spatial-aware Denoising Network for Open-vocabulary Panoptic Scene Graph Generation with Long- and Local-range Context Reasoning},
  booktitle = {Proceedings of the IEEE/CVF International Conference on Computer Vision},
  year = {2025}
}

@inproceedings{yin2025knowledge,
  author = {Yin, W. and Wang, Y. and Duan, G. and Zhang, D. and Hu, X. and Li, Y.-F. and He, T.},
  title = {Knowledge-aligned Counterfactual-enhancement Diffusion Perception for Unsupervised Cross-domain Visual Emotion Recognition},
  booktitle = {Proceedings of the IEEE/CVF Conference on Computer Vision and Pattern Recognition},
  pages = {3888--3898},
  year = {2025}
}

@inproceedings{owusu2024graph,
  author = {Owusu, J. W. and Zakari, R. Y. and Qin, K. and He, T.},
  title = {Graph Convolutional Networks with Fine-Tuned Word Representations for Visual Question Answering},
  booktitle = {2024 IEEE Smart World Congress},
  pages = {1381--1387},
  year = {2024}
}

@inproceedings{yang2024towards,
  author = {Yang, Z. and Liu, X. and Ouyang, D. and Duan, G. and Zhang, D. and He, T. and Li, Y.-F.},
  title = {Towards Open-vocabulary HOI Detection with Calibrated Vision-language Models and Locality-aware Queries},
  booktitle = {Proceedings of the 32nd ACM International Conference on Multimedia},
  pages = {1495--1504},
  year = {2024}
}

@article{he2024towardslifelong,
  author = {He, T. and Wu, T. and Zhang, D. and Duan, G. and Qin, K. and Li, Y.-F.},
  title = {Towards Lifelong Scene Graph Generation with Knowledge-aware In-context Prompt Learning},
  journal = {arXiv preprint arXiv:2401.14626},
  year = {2024}
}

@article{he2023toward,
  author = {He, T. and Gao, L. and Song, J. and Li, Y.-F.},
  title = {Toward a Unified Transformer-based Framework for Scene Graph Generation and Human-Object Interaction Detection},
  journal = {IEEE Transactions on Image Processing},
  volume = {32},
  pages = {6274--6288},
  year = {2023}
}

@article{he2022state,
  author = {He, T. and Gao, L. and Song, J. and Li, Y.-F.},
  title = {State-aware Compositional Learning Toward Unbiased Training for Scene Graph Generation},
  journal = {IEEE Transactions on Image Processing},
  volume = {32},
  pages = {43--56},
  year = {2022}
}

@inproceedings{he2022towards,
  author = {He, T. and Gao, L. and Song, J. and Li, Y.-F.},
  title = {Towards Open-vocabulary Scene Graph Generation with Prompt-based Finetuning},
  booktitle = {European Conference on Computer Vision},
  year = {2022}
}

@article{he2021semantic,
  author = {He, T. and Gao, L. and Song, J. and Cai, J. and Li, Y.-F.},
  title = {Semantic Compositional Learning for Low-shot Scene Graph Generation},
  journal = {arXiv preprint arXiv:2108.08600},
  year = {2021}
}

@inproceedings{he2021exploiting,
  author = {He, T. and Gao, L. and Song, J. and Li, Y.-F.},
  title = {Exploiting Scene Graphs for Human-Object Interaction Detection},
  booktitle = {Proceedings of the IEEE/CVF International Conference on Computer Vision},
  pages = {15984--15993},
  year = {2021}
}

@inproceedings{he2020learning,
  author = {He, T. and Gao, L. and Song, J. and Cai, J. and Li, Y.-F.},
  title = {Learning from the Scene and Borrowing from the Rich: Tackling the Long Tail in Scene Graph Generation},
  booktitle = {Proceedings of the International Joint Conference on Artificial Intelligence},
  year = {2020}
}

@article{zakari2025vqa,
  author = {Zakari, R. Y. and Owusu, J. W. and Qin, K. and Wang, H. and Lawal, Z. K. and He, T.},
  title = {VQA and Visual Reasoning: An Overview of Approaches, Datasets, and Future Direction},
  journal = {Neurocomputing},
  volume = {622},
  pages = {129345},
  year = {2025}
}

@inproceedings{song2017deepdiscrete,
  author = {Song, J. and He, T. and Fan, H. and Gao, L.},
  title = {Deep Discrete Hashing with Self-supervised Pairwise Labels},
  booktitle = {Joint European Conference on Machine Learning and Knowledge Discovery in Databases},
  year = {2017}
}

@inproceedings{he2020sneq,
  author = {He, T. and Gao, L. and Song, J. and Wang, X. and Huang, K. and Li, Y.},
  title = {SNEQ: Semi-supervised Attributed Network Embedding with Attention-based Quantisation},
  booktitle = {Proceedings of the AAAI Conference on Artificial Intelligence},
  volume = {34},
  pages = {4091--4098},
  year = {2020}
}

@article{he2021semisupervised,
  author = {He, T. and Gao, L. and Song, J. and Li, Y.-F.},
  title = {Semisupervised Network Embedding with Differentiable Deep Quantization},
  journal = {IEEE Transactions on Neural Networks and Learning Systems},
  volume = {34},
  number = {8},
  pages = {4791--4802},
  year = {2021}
}

@article{dai2024muap,
  author = {Dai, R. and Tan, Y. and Mo, L. and He, T. and Qin, K. and Liang, S.},
  title = {MUAP: Multi-step Adaptive Prompt Learning for Vision-Language Model with Missing Modality},
  journal = {arXiv preprint arXiv:2409.04693},
  year = {2024}
}

@article{zhang2024cviformer,
  author = {Zhang, D. and Liang, S. and He, T. and Shao, J. and Qin, K.},
  title = {CVIformer: Cross-View Interactive Transformer for Efficient Stereoscopic Image Super-Resolution},
  journal = {IEEE Transactions on Emerging Topics in Computational Intelligence},
  volume = {9},
  number = {2},
  year = {2024}
}
}

\end{document}